\RequirePackage{amsmath}
\documentclass[runningheads]{llncs}
\usepackage[T1]{fontenc}
\usepackage{amsfonts}
\usepackage{bm}
\usepackage{lineno}
\usepackage{tabularx}
\usepackage{multirow}
\usepackage{multicol}
\usepackage{subfig}
\usepackage{graphicx}
\usepackage{tikz}
\usepackage{color}
\usepackage{xspace}
\usepackage{enumerate}
\usepackage{xurl}
\usepackage[colorlinks,allcolors=blue]{hyperref}
\usepackage{xspace}
\usepackage[normalem]{ulem}
\usepackage[acronym, shortcuts]{glossaries}
\usepackage{colortbl}

\newcolumntype{L}{>{\raggedright\arraybackslash}X}
\newcolumntype{C}{>{\centering\arraybackslash}X}
\newcolumntype{R}{>{\raggedleft\arraybackslash}X}

\newcommand{\mat}[1]{\mbox{\fontencoding{T1}\sffamily\slshape{#1\/}}} 
\newcommand{\func}[1]{{\mbox{\usefont{OT1}{pzc}{m}{it}{#1}}}}
\newcommand{\set}[1]{\mathcal{#1}}

\newcommand\rurl[1]{%
  \href{https://#1}{\nolinkurl{#1}}%
}

\newacronym{eo}{EO}{Earth Observation}
\newacronym{ml}{ML}{Machine Learning}
\newacronym{dl}{DL}{Deep Learning}
\newacronym{sits}{SITS}{Satellite Image Time Series}
\newacronym{cca}{CCA}{Canonical Correlation Analysis}
\newacronym{f1}{F1}{F1 Macro}
\newacronym{r2}{$R^2$}{Coefficient of Determination}
\newacronym{prs}{PRS}{Performance Robustness Score}
\newacronym{lfmc}{LFMC}{Live Fuel Moisture Content}
\newacronym{pm25}{PM25}{Particulate Matter 2.5}
\newacronym{cropharvest}{CH}{CropHarvest}
\newacronym{mlp}{MLP}{Multi-Layer Perceptron}
\newacronym{gru}{GRU}{Gated Recurrent Unit}
\newacronym{maug}{MAUG}{Missing as AUGmentation}
\newacronym{td}{TempD}{Temporal Dropout} 
\newacronym{sd}{SensD}{Sensor Dropout} 
\newacronym{isd}{ISensD}{Input Sensor Dropout}
\newacronym{sensei}{ESensI}{Ensemble Sensor Invariant}

\begin{document}
\title{Increasing the Robustness of Model Predictions to Missing Sensors in Earth Observation}
\titlerunning{Increasing Robustness of Multi-sensor Model to Missing Sensors}

\author{Francisco Mena\inst{1,2}
\and
Diego Arenas\inst{2}
\and
Andreas Dengel\inst{1,2}
}
\authorrunning{F. Mena et al.}

\institute{University of Kaiserslautern-Landau, Kaiserslautern, Germany \\
German Research Center for Artificial Intelligence, Kaiserslautern, Germany\\
\email{f.menat@rptu.de}
}

\maketitle              
\begin{abstract} 
Multi-sensor ML models for EO aim to enhance prediction accuracy by integrating data from various sources. 
However, the presence of missing data poses a significant challenge, particularly in non-persistent sensors that can be affected by external factors. 
Existing literature has explored strategies like temporal dropout and sensor-invariant models to address the generalization to missing data issues. 
Inspired by these works, we study two novel methods tailored for multi-sensor scenarios, namely Input Sensor Dropout (ISensD) and Ensemble Sensor Invariant (ESensI). 
Through experimentation on three multi-sensor temporal EO datasets, we demonstrate that these methods effectively increase the robustness of model predictions to missing sensors. 
Particularly, we focus on how the predictive performance of models drops when sensors are missing at different levels.
We observe that ensemble multi-sensor models are the most robust to the lack of sensors. In addition, the sensor dropout component in ISensD shows promising robustness results.
\end{abstract}

\section{Introduction} \label{sec:introduction}

Multi-sensor \gls{ml} models for \gls{eo} use data from diverse sources, providing a comprehensive view of the Earth. These sources can be from ground-based to satellite-based sensors. Recently, the literature has focused on using these multi-sensor models to improve accuracy regarding single-sensor models \cite{mena2024common}. 
However, by including more sensors, there is a greater chance that one can face technical issues and miss some data \cite{little2019statistical}.

Missing data is inherent in the \gls{eo} domain, hindering accurate predictions and introducing biases.
Moreover, \gls{ml} models, even adaptive models such as transformers, are not naturally robust to missing data \cite{ma2022multimodal}. 
In \gls{sits}, cloudy conditions hinder the availability of optical images, negatively affecting model predictions \cite{saintefaregarnot2022multi}.
The decline in performance has also been observed when all features of a specific sensor are unavailable in land-use classification \cite{hong2021more} and vegetation applications \cite{mena2024igarss}.

The design of \gls{ml} models aligned to the missing data scenario has been vaguely explored in the literature. 
Garnot et al. \cite{saintefaregarnot2022multi} present a \gls{td} technique that randomly drops some time-steps in \gls{sits} data, obtaining a more robust model to missing time-steps. However, the \gls{td} is focused on single-sensor input \cite{saintefaregarnot2022multi}.
On the other hand, Francis et al. \cite{francis2024sensor} proposed a sensor invariant model that can yield predictions with similar accuracy, regardless of the sensor given as input. However, the proposed model focuses on just static images.
Motivated by these works, we introduce two prediction methods tailored to the multi-sensor time-series scenario.

The main research questions of our work are: how to handle missing sensors in models at prediction time? how to increase the prediction robustness of model to missing sensors?
We use three multi-sensor time series \gls{eo} datasets to validate our methods in scenarios of missing sensor data. In addition, we show that we can increase the model robustness compared to four methods from the literature.
\section{Handling missing data in multi-sensor learning} \label{sec:basis}

Multi-sensor learning considers the case when multiple set of features (from different sensors) are used as input in predictive \gls{ml} models. 
The main objective is to corroborate the information on individual observations \cite{mena2024common}. The \gls{eo} literature evidence that using a multi-sensor perspective is crucial to enrich the input data and improve model performance \cite{hong2021more,saintefaregarnot2022multi,mena2024search}. 
However, an appropriate modeling is necessary to get the most advantage of the multi-sensor data \cite{ma2022multimodal,pathak2023predicting,mena2024common}. 

When multi-sensor \gls{ml} models are deployed in real environments, the assumption that sensor data is persistently available cannot be taken. During inference, the occurrence of missing sensors leads to a scenario for which the model is unprepared.  
Since re-training can be impractical due to limited resources, some techniques can help to handle in a better way the missing of sensor data.
In the following, we describe such techniques commonly used in the literature.

\paragraph{Impute.}
As \gls{ml} models operate with fixed-size matrices, if sensor features are missing, we need to fill in some numerical value to yield a prediction. 
This has been studied with \gls{eo} data when all features of a specific sensor are missing. For instance, when optical or radar images are missing in land-cover classification \cite{hong2021more}, and when optical or radar time series are missing in different vegetation applications \cite{mena2024igarss}.
The single hyperparameter of this technique is the imputation value. For instance, Srivastava et al. \cite{srivastava2012multimodal} used random values, Hong et al. \cite{hong2021more} used zero, and Mena et al. \cite{mena2024igarss} the average of each feature in the training data.

\paragraph{Exemplar.}
Another option to filling in missing sensor features is to use existing values observed during training. 
Then, the missing features can be replaced with a similar sample via a training-set lookup. For instance, Srivastava et al. \cite{srivastava2019understanding} performs the search with the sensors available in a shared space. The shared space is obtained with a linear \gls{cca} projection from features learned by sensor-dedicated encoders. Aksoy et al. \cite{aksoy2009land} uses a $k$-nearest neighbor algorithm to search for a candidate value to replace the missing features.
The hyperparameters of this technique are how to define the shared space, which similarity function to use, and which sensors to use in the search.

\paragraph{Reconstruction.}
A more sophisticated way to fill in the missing sensors is to learn a function that predicts all the missing sensor features. 
Such techniques include cross-modal auto-encoders and translation models \cite{srivastava2012multimodal}. 
The latter method has been applied to \gls{eo} data by translating a radar to optical image with cycleGAN \cite{efremova2021soil}, or using multiple optical sensors at different times to reconstruct the optical features at a specific time \cite{scarpa2018cnn}.  
Here, the practitioner has to define the entire reconstruction model, as well as its optimization.

\paragraph{Ignore.}
A multi-sensor \gls{ml} model can be designed in a way to explicitly ignore the missing sensors. 
For instance, in the dynamic ensemble framework \cite{ko2008dynamic} the models used in the aggregation are dynamically selected for each sample, completing ignoring the predictions of other models. Recently, this technique was applied to \gls{eo} data where a sensor-dedicated model was trained on each sensor \cite{mena2024igarss}. In this case, the predictions from the missing sensors are simply omitted in the aggregation.
Furthermore, it has been applied to average intermediate features in the model by ignoring sensor-dedicated features \cite{mena2024igarss}.

While prior methods solely address the handling of missing sensor data, our work goes beyond this issue by introducing methods aimed at enhancing the model's robustness under missing sensor data.
\section{Methods} \label{sec:methods}

Two multi-sensor \gls{ml} models are proposed to tackle the missing sensor problem.

\subsection{\gls{isd}} \label{sec:methods:isd}
\begin{figure}[!t]
    \centering
    \includegraphics[width=0.73\textwidth]{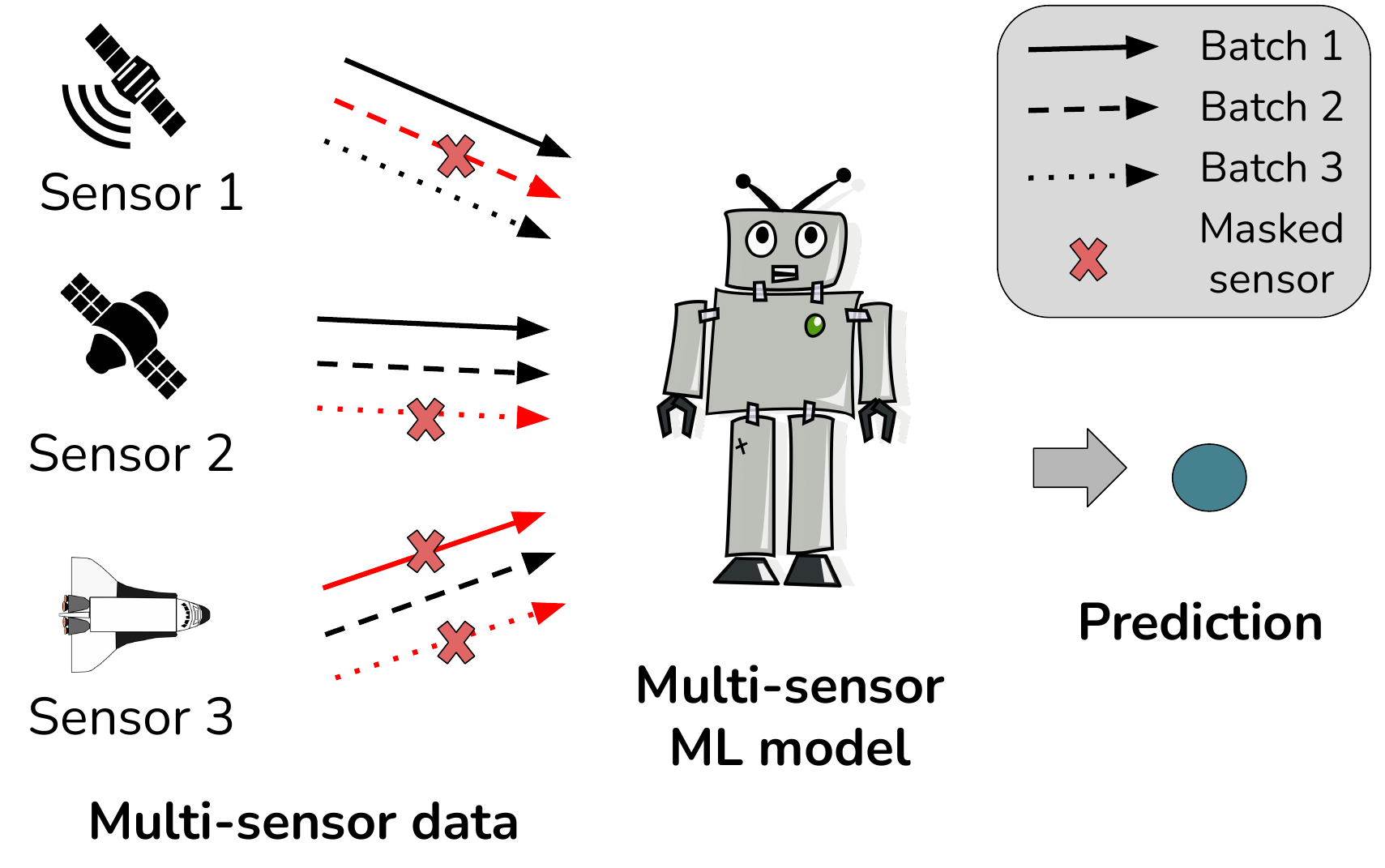}
    \caption{Illustration of the \acrshort{isd} method. We show three batch examples and the associated randomly masked sensors based on the \acrshort{sd} technique.} \label{fig:sensordropout}
\end{figure}
Inspired by recent works that use dropout as data augmentation \cite{saintefaregarnot2022multi}, we introduce a \gls{sd} technique at the input-level. The idea is to randomly mask out all input features of some sensors during training. 
Consider the set of sensor features (input data) for a sample $i$ as $\set{X}^{(i)} = \{ \mat{X}^{(i)}_s \}_{s \in \set{S}}$, with $\set{S}$ the set of all available sensors. Then, the masked input features can be expressed by $\tilde{\set{X}}^{(i)} =\{ d_s^{(i)} \cdot \mat{X}_s^{(i)} \}_{s \in \set{S}} $, with $d_s^{(i)} \sim \text{Bernoulli}(r)$ and $r \in [0,1]$ the \gls{sd} ratio. This ratio that controls how much the sensor features will be masked out. 
In addition, we introduce a special case of the previous that does not rely on the hyperparameter $r$. The alternative masking technique lists all the possible combinations of missing sensors ($2^{ | \set{S} |} - 1 $) and then selects one randomly, which we called No Ratio (NR). 
We apply these techniques in a batch-wise manner, with a different random masking $d_s^{(i)}$ at each batch. 
An illustration of the masking is shown in Fig.~\ref{fig:sensordropout}. 
Finally, the masked features are given as input to a multi-sensor model implementing input-level fusion \cite{mena2024common}, i.e. $\hat{y}^{(i)} = \func{M}_{\theta}( \func{concat} ( \tilde{\set{X}}^{(i)}))$, with $\func{M}_{\theta}$ the model function parameterized by $\theta$. 
In Sec.~\ref{sec:experiments} we present the model used in $\func{M}_{\theta}$.
Finally, the parameters of the model are optimized regarding the ground truth $y^{(i)}$ in the training set by
\begin{equation}
   \hat{\theta} =  \arg\min_{\theta} \sum_{i=1}^N \func{L} (y^{(i)}, \hat{y}^{(i)}) \ .
\end{equation}

\subsection{\gls{sensei}} \label{sec:methods:sensei}
\begin{figure}[!t]
    \centering
    \includegraphics[width=0.75\textwidth]{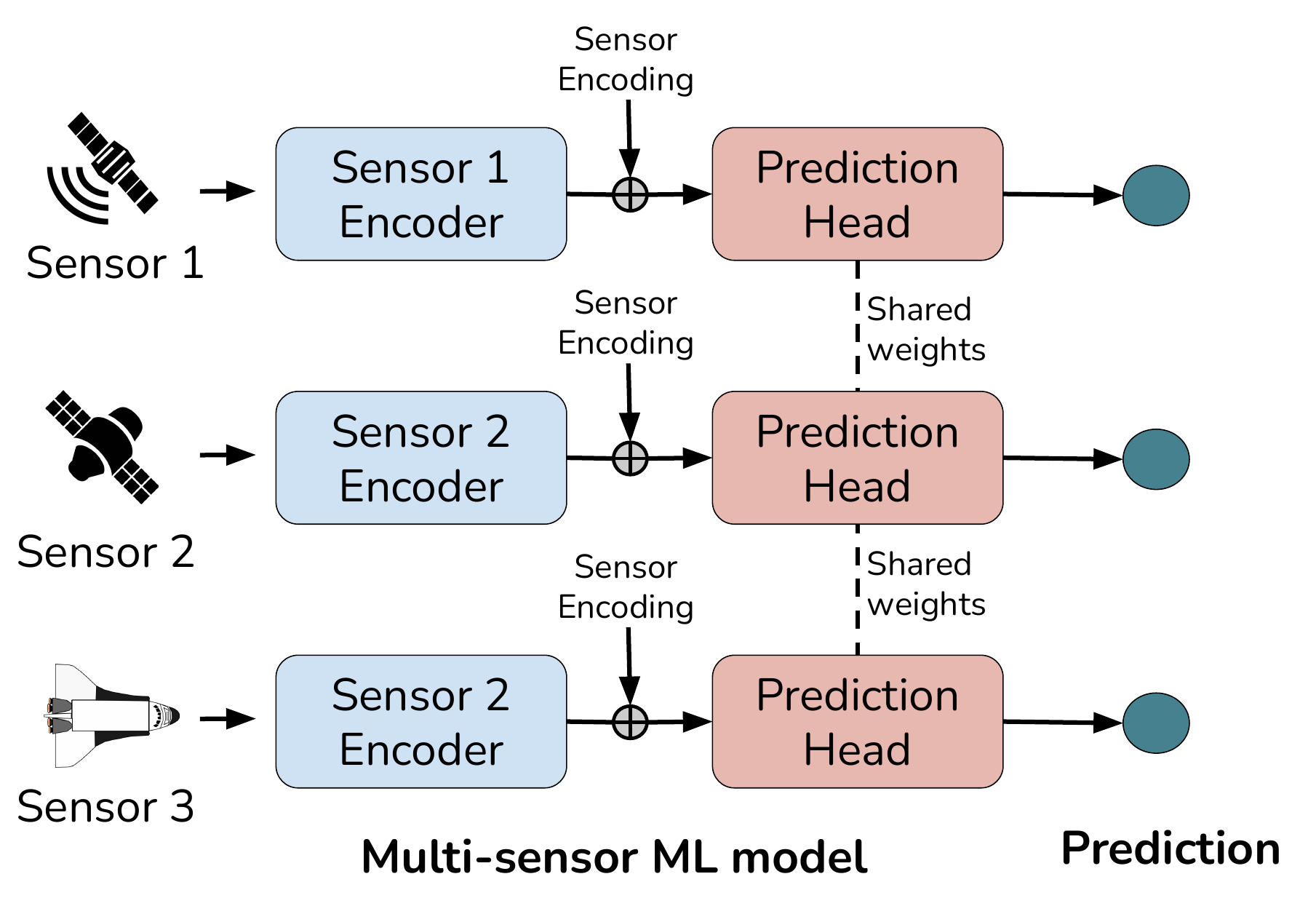}
    \caption{Illustration of the \acrshort{sensei} method. The main component is the shared prediction head, which makes the ensemble-based a sensor-invariant model.} \label{fig:sensorinvariant}
\end{figure}

Inspired by recent works that use sensor invariant models in the context of \gls{eo} \cite{francis2024sensor}, we propose a model with sensor invariant layers. The idea is to use an ensemble multi-sensor model with a sensor invariant prediction head. 
Consider the standard ensemble-based model with sensor-dedicated encoders $\vec{z}_s^{(i)} = \func{E}_{\theta_s}( \mat{X}_s^{(i)}) \in \mathbb{R}^d$, and sensor-dedicated prediction heads $\hat{y}_s^{(i)} = \func{P}_{\phi_s}(\vec{z}_s^{(i)})$. Then, the parameters of each sensor-dedicated model $\Theta_s = \{ \theta_s, \phi_s \}$ are optimized by a loss function $\mathcal{L}$
\begin{equation}
   \hat{\Theta}_s =  \arg\min_{\Theta_s} \sum_{i=1}^N \func{L} (y^{(i)}, \hat{y}_s^{(i)}) \ .
\end{equation}
Since these sensor-dedicated parameters $\Theta_s$ are optimized independently of each other, we incorporate the notion of multi-sensor information by using a shared prediction head in all sensors: $\hat{y}_s^{(i)} = \func{P}_{\phi}(\vec{z}_s^{(i)} + \vec{\rho}_s) \ \forall s \in \mathcal{S}$. We use a learnable sensor encoding vector $\vec{\rho}_s \in \mathbb{R}^d$ inspired by the channel group encoding in \cite{tseng2023lightweight}. The purpose of the sensor encoding vector is to give the model a notion of awareness of which sensor features are being fed for prediction. An illustration of this method is shown in Fig.~\ref{fig:sensorinvariant}. Since the features $z_s$ can have quite different magnitudes, we consider the case in which $\func{P}_{\phi}$ incorporates a normalization in its first layer. Then, as the parameters of the sensor-dedicated model $\Theta_s = \{ \theta_s, \rho_s, \phi \}$ are attached by $\phi$ (the shared prediction head), these are jointly optimized as
\begin{equation}
   \hat{\Theta} =  \arg\min_{ \left\{ \Theta_s\right\}_{s\in \set{S}}  } \sum_{i=1}^N \func{L} (y^{(i)}, \hat{y}_s^{(i)}) \ .
\end{equation}

\section{Experiments} \label{sec:experiments}

\subsection{Datasets}

\paragraph{\Gls{cropharvest}.} 
We use the cropharvest dataset for multi-sensor crop recognition \cite{tseng2021crop}, in a multi-crop version \cite{mena2024igarss}.
This involves a classification task in which the crop-type (between 10 crop-type groups including \textit{non-crop}) at a given location during a particular season is predicted.
The dataset has 29,642 samples around the globe between 2016 and 2022. 
The temporal sensors are multi-spectral optical (from Sentinel-2), radar (from Sentinel-1), and weather (from ERA5).
These were re-sampled monthly during the crop growing year. 
An additional static sensor is the topographic information (from SRTM's DEM).
All sensor features are spatially interpolated to a pixel resolution of 10 m.

\paragraph{\Gls{lfmc}.}
We use a dataset for multi-sensor moisture content estimation \cite{rao2020sar}. 
This involves a regression task in which the vegetation water (moisture) per dry biomass (in percentage) in a given location at a specific time is predicted.
There are 2,578 samples from the western US collected between 2015 and 2019.
The temporal sensors are multi-spectral optical (from Landsat-8) and radar (from Sentinel-1). 
These features were re-sampled monthly during a four-month window before the moisture measurement. 
Additional static sensors are the topographic information (from NED), 
soil properties (from Unified North American Soil Map), 
canopy height (from GLAS's lidar), 
and land-cover class (from GlobCover).
All features are interpolated to a pixel resolution of 250 m.

\paragraph{\Gls{pm25}.}
We use a dataset for multi-sensor air quality estimation \cite{pm25}. 
This involves a regression task in which the concentration (in $ug/m^3$) of PM2.5 in the air (particles that are 2.5 microns or less in diameter) in a city at a specific moment is predicted.
There are 167,309 samples in five Chinese cities between 2010 and 2015.
The temporal sensor features are atmospheric conditions, atmospheric dynamics, and precipitation.
These sensors are captured at hourly resolution. We consider a three-day window of information before the day of the PM2.5 measurement.

\subsection{Baselines}
We consider four baselines in the \gls{eo} literature for missing sensors.

\begin{itemize}
    \item Input: This corresponds to a standard input-level fusion model commonly used in the literature. For the case of missing sensors, the missing input features are replaced by the mean of the features in the training set \cite{mena2024igarss}.
    \item ITempD: This corresponds to the Input method in addition to using the \gls{td} technique. For the case of missing sensors, the missing input features are replaced by the same masking value used in the \gls{td} \cite{saintefaregarnot2022multi}.
    \item Feature: This corresponds to a feature level fusion model that concatenates the features extracted by sensor-dedicated encoders. For the case of missing sensors, the missing features are replaced with the exemplar technique (similarity-based search) proposed in \cite{srivastava2019understanding}.
    \item Ensemble: This corresponds to an ensemble-based aggregation model that averages the predictions of sensor-dedicated models. For the case of missing sensors, the missing predictions are just ignored during the aggregation \cite{mena2024igarss}.
\end{itemize}
For the input-level fusion methods, we align the features of the temporal and static sensors by repeating the static features across the time series, as commonly done in \gls{eo} \cite{tseng2021crop,saintefaregarnot2022multi,pathak2023predicting}.
Besides, for the model $\func{M}_{\theta}$, and $\func{E}_{\theta}$ for temporal features (Sec.~\ref{sec:methods}) we use a 1D CNN. We show results with alternative model architectures in the appendix. For the prediction head $\func{P}_{\phi}$, and static features (in Feature and Ensemble), we use an MLP encoder. Two layers with 128 units and a 20\% of dropout are used.
We use the ADAM optimizer with a batch-size of 128 over the cross-entropy in classification and mean squared error in regression. 

\subsection{Experimental setting}

We assess the methods with a 10-fold cross-validation by simulating missing sensors in the validation fold. The assessment consists of having fewer sensors at validation time than during training to make the prediction.
We experiment with different degrees of missingness controlled by a percentage of samples that are affected by missing sensors. 
We consider that sensors with temporal features might be missing, leaving the static sensors as permanently available information. 
For evaluating the model robustness, we use the \gls{prs} proposed in \cite{heinrich2023targeted}. 
This metric is based on the predictive error with missing sensors relative to the predictive error when using all sensors: 
\begin{equation}
\text{PRS}(y, \hat{y}_{miss}, \hat{y}_{full}) = \min \left(1, \ \exp{ \left( 1- \frac{\text{RMSE}(y, \hat{y}_{miss})}{ \text{RMSE}(y, \hat{y}_{full})} \right) } \right)    
\end{equation}
For the predictive performance, we use the \gls{f1} macro in classification and the \gls{r2} in regression tasks.

\subsection{Results}\label{sec:experiments:resuts}

We report the results of the best dropout ratio in the \gls{sd}, which is 60\% in \gls{cropharvest} data, 25\% in the \gls{lfmc} data, and 45\% in the \gls{pm25} data. For the \gls{sensei} we include the sensor encoding vector in \gls{cropharvest} data, in \gls{lfmc} we add the sensor encoding with normalized vectors, and in \gls{pm25} the vector is not included. 
Ablation studies on these parameters are shown in Sec.~\ref{sec:experiments:abl}.

\begin{figure}[!t]
    \centering
    \includegraphics[width=0.99\textwidth]{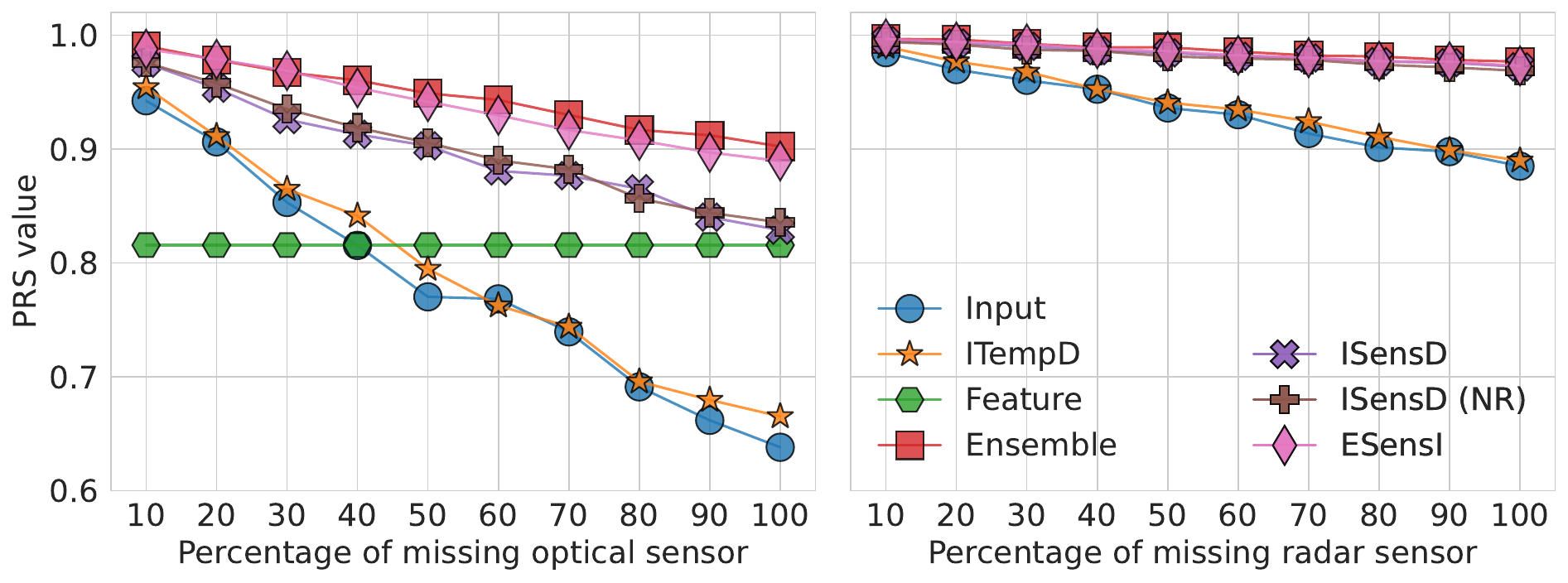}
    \caption{\gls{prs} of methods at different levels of missing sensors in the \gls{cropharvest} data.} \label{fig:cropharvest_prs_res}
\end{figure}

\begin{figure}[!t]
    \centering
    \includegraphics[width=0.99\textwidth]{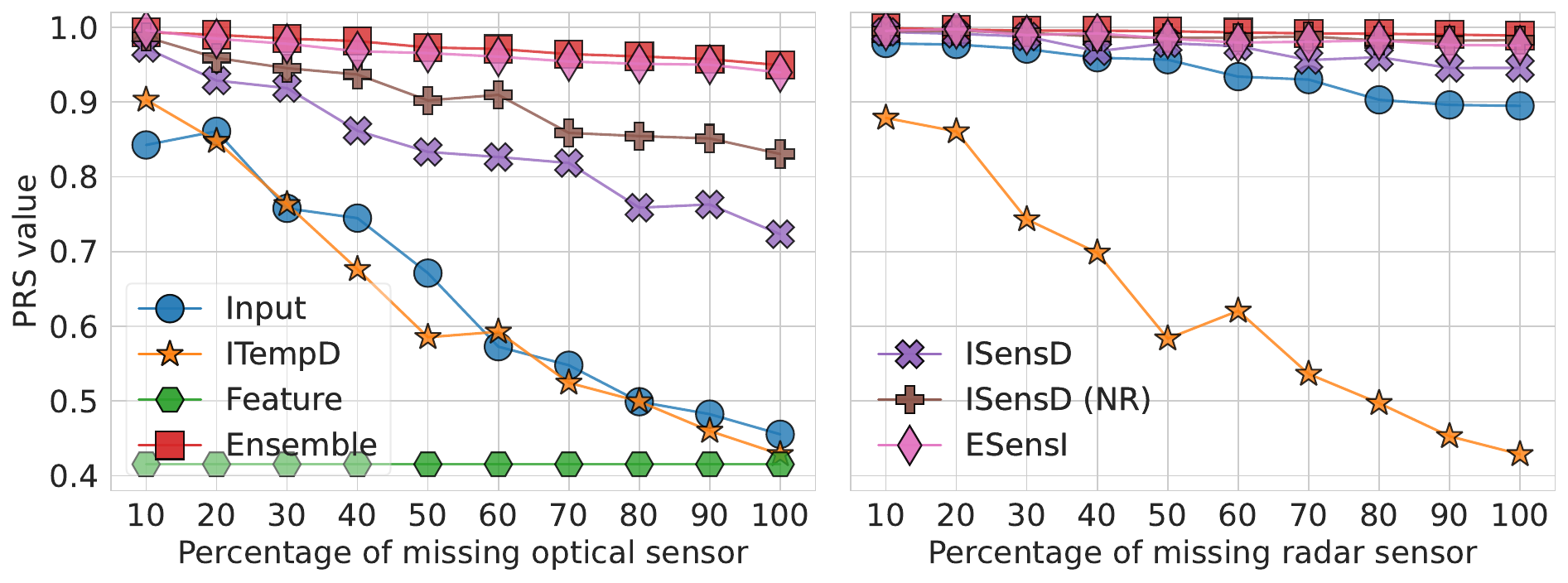}
    \caption{\gls{prs} of methods at different levels of missing sensors in the \gls{lfmc} data.} \label{fig:lfmc_prs_res}
\end{figure}

\begin{figure}[!t]
    \centering
    \includegraphics[width=0.99\textwidth]{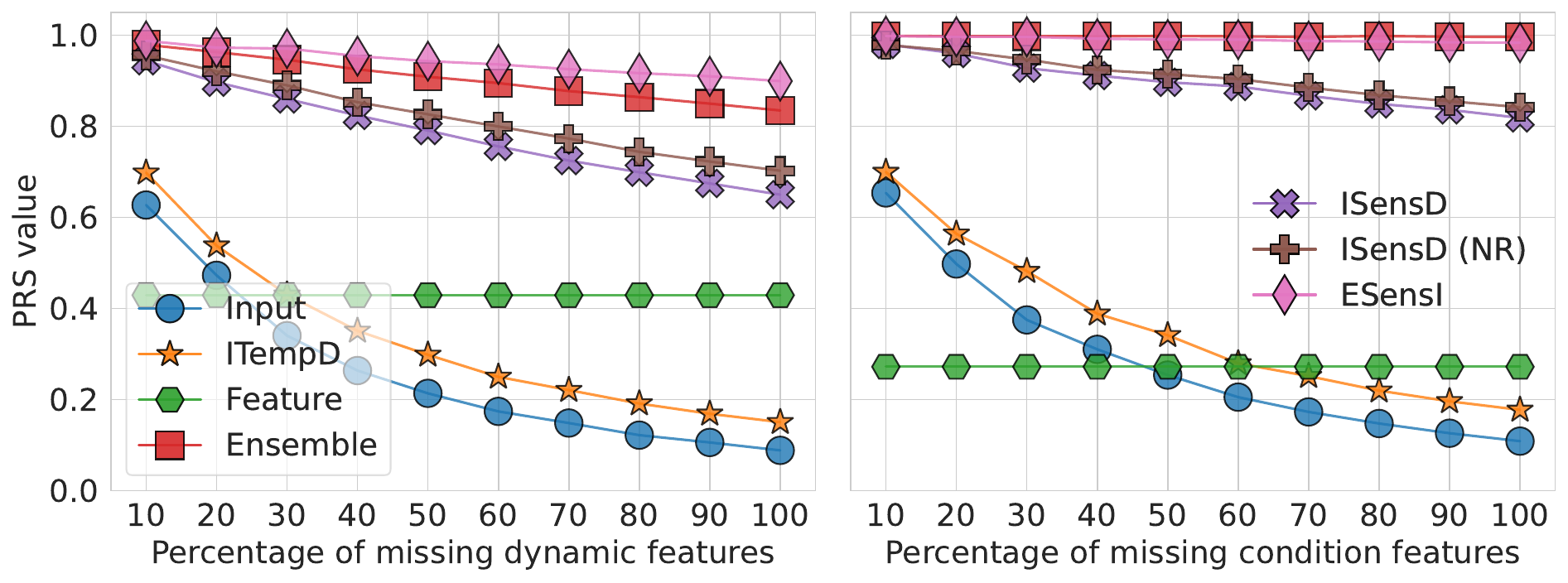}
    \caption{\gls{prs} of methods at different levels of missing sensors in the \gls{pm25} data.} \label{fig:pm25_prs_res}
\end{figure}

In Fig~\ref{fig:cropharvest_prs_res}, \ref{fig:lfmc_prs_res}, and \ref{fig:pm25_prs_res} we display the PRS of the compared methods in the \gls{cropharvest}, \gls{lfmc}, and \gls{pm25} data respectively. 
Overall, we observe that Ensemble and \gls{sensei} methods are the most robust when increasing the level of missing sensors in the datasets.
In addition, Input and ITempD methods have a linear decrease in the robustness with different slopes, depending on the dataset and which sensor is missing. For instance, in the \gls{cropharvest} and \gls{lfmc} data, Fig~\ref{fig:cropharvest_prs_res} and \ref{fig:lfmc_prs_res}, the decrease is higher when the optical sensor is missing. 
Besides, in the \gls{pm25} data, these methods have an exponential decrease in robustness.

The \gls{prs} results highlight that, by increasing the percentage of missing sensor data, our proposals show stronger robustness (\gls{prs} decreases by a smaller amount) in all cases.
The \gls{isd} robustness is significantly increased regarding its direct baselines, i.e. Input and ITempD methods. 
However, the best robustness is achieved by the Ensemble method, which in some cases (Fig.~\ref{fig:pm25_prs_res}) is improved by the \gls{sensei} method.
This illustrates that the proposed methods can increase the robustness of multi-sensor models.
The same pattern is observed in the predictive performance results (with \gls{f1} and \gls{r2}), presented in the appendix.

\begin{table}[!t]
    \centering
    \caption{Predictive performance on a full-sensor evaluation, i.e. no missing features, on different datasets. The \acrshort{f1} is shown in classification and \acrshort{r2} in regression.}     \label{tab:nomissing_results}
    \begin{tabularx}{\textwidth}{c|CCCC|CcC} \hline
         Dataset & Input & ITempD & Feature & Ensemble & \gls{isd} & \gls{isd} (NR) & \gls{sensei}  \\ \hline
         \gls{cropharvest}  &  $\mathbf{0.648}$ & $0.642$ & $0.576$ & $0.605$ & $0.598$ & $0.611$ & $0.615$ \\
         \gls{lfmc} & $0.714$ & $\mathbf{0.717}$ & $0.650$ & $0.313$ & $0.655$ & $0.545$ & $0.326$ \\
         \gls{pm25} & $\mathbf{0.917}$ & $0.882$ & $-0.354$  & $0.308$ & $0.525$ & $0.496$ & $0.232$ \\ \hline
    \end{tabularx}
\end{table}
In general, the PRS results indicate that the Ensemble method is quite robust without the sensor invariant component, i.e. \gls{sensei}. 
However, when we analyze the predictive performance with all sensors available, in Table~\ref{tab:nomissing_results}, it reflects that the Ensemble method is among the worst in predictive performance, with the Input method at the top. 
Here, we observe that the \gls{sensei} method improves in two out of three datasets in comparison to the Ensemble method. 
However, the \gls{isd} method performs worse than its counterpart, Input, reflecting the limitation of the \gls{sd} technique when all sensors are available.
We suspect that the model is focusing more on the missing data, ignoring the full-sensor scenario, as this occurs only once within all the randomized scenarios.

\subsection{Ablation study}\label{sec:experiments:abl}

\begin{figure}[!t]
    \centering
    \subfloat
    {\includegraphics[width=0.9\textwidth]{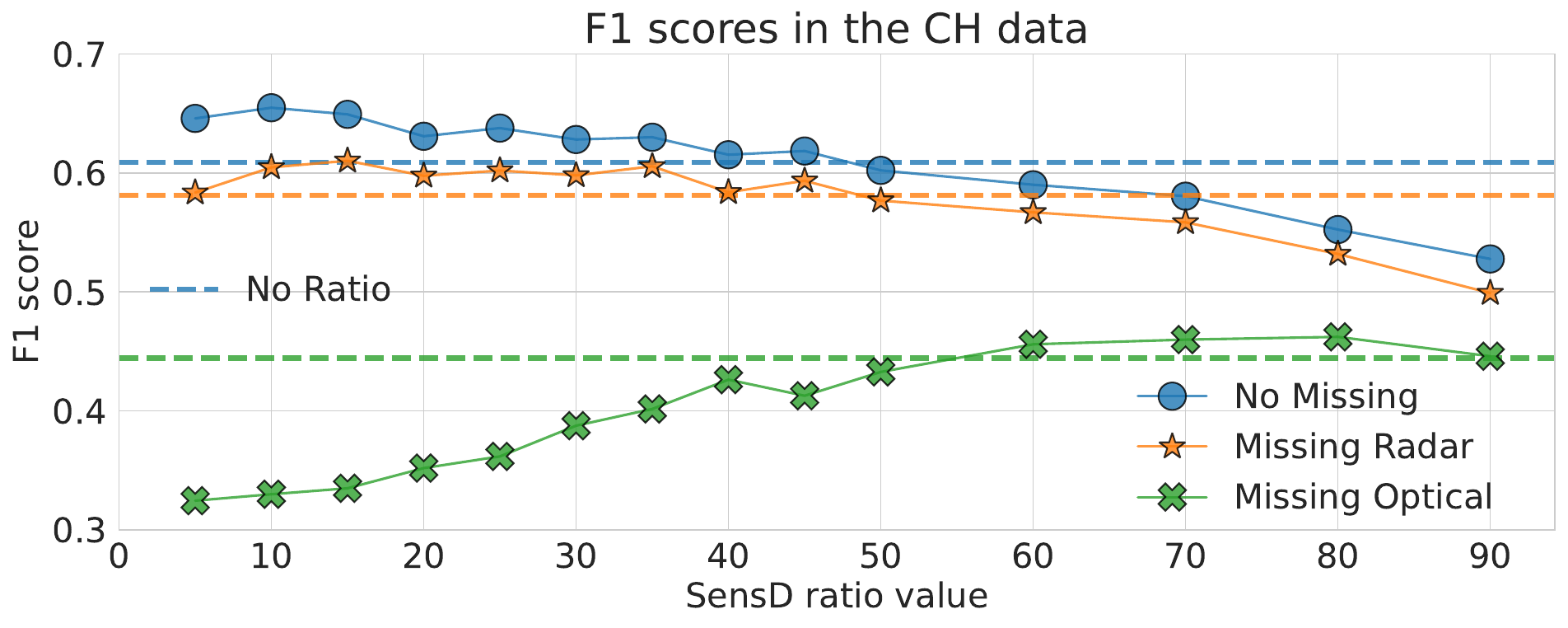}} \\
    \subfloat
    {\includegraphics[width=0.9\textwidth]{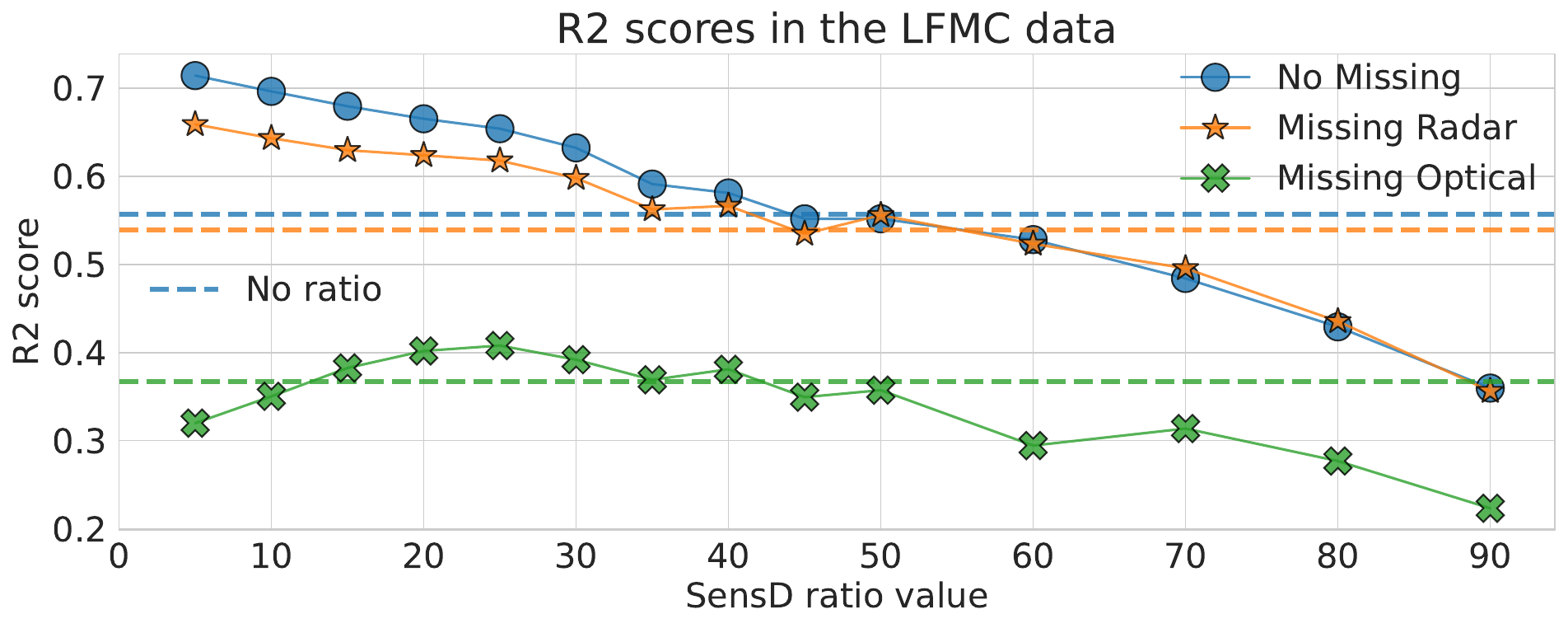}}\\
    \subfloat
    {\includegraphics[width=0.9\textwidth]{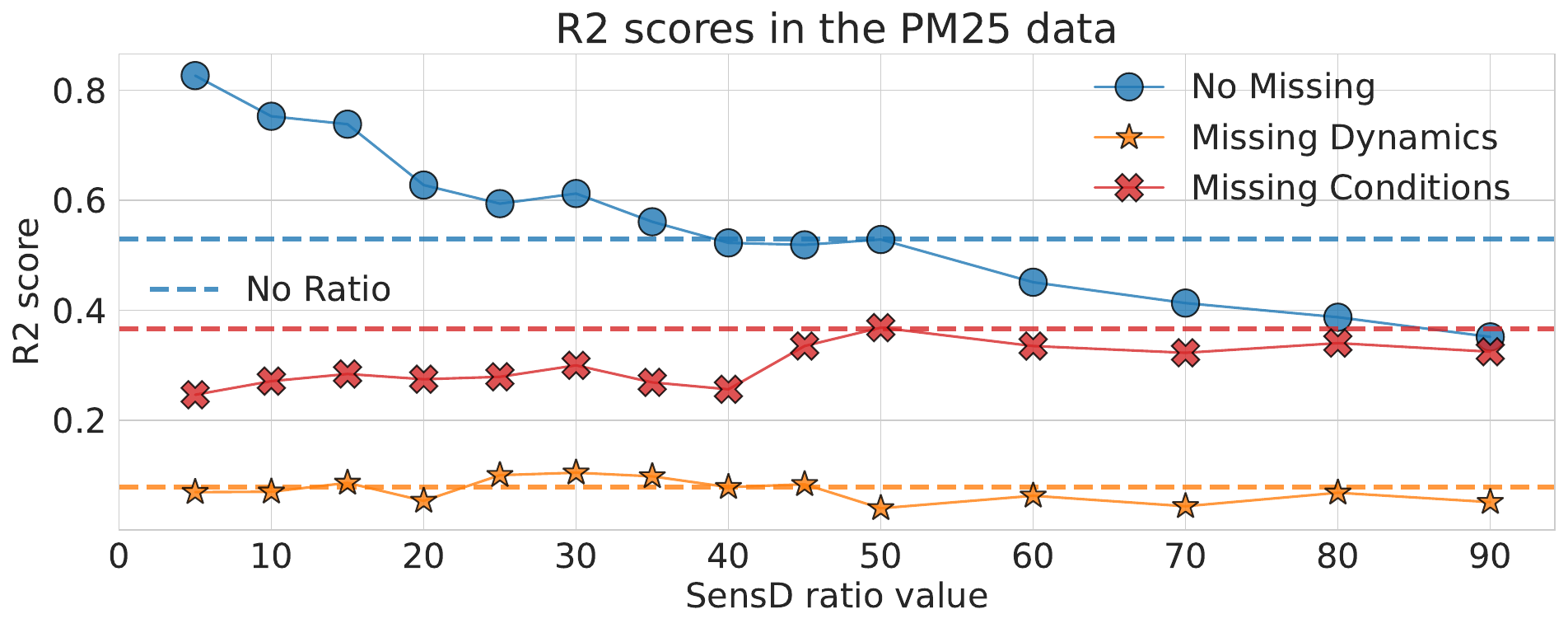}}
    \caption{Predictive performance results of the \gls{isd} method with different \gls{sd} ratio values. The dashed lines show the no ratio version of the \gls{isd} method.} \label{fig:drop_ratio}
\end{figure}

We assess different configurations in the methods presented.
Fig.~\ref{fig:drop_ratio} displays the predictive performance when different values of the \gls{sd} ratio are used in the \gls{isd} method. 
Depending on the missing sensor, increasing the ratio increases or decreases the performance, while for the full-sensor scenario, it always tends to decrease.
However, the results with different \gls{sd} ratios variate around the NR version, showing its good capacity as a parameter-free alternative.
For the \gls{sensei} method, we show the predictive performance with different model components in Table~\ref{tab:sensei_config}. 
In this case, it reflects that simply sharing weights is not optimal in the \gls{cropharvest} and \gls{lfmc} data, but rather additional components should be added. On the contrary, in \gls{pm25}, it is better to use shared weights in the model prediction head.

\begin{table}[!t]
    \centering
    \caption{Predictive performance results of different configurations in the \gls{sensei} model.  The results are obtained in the full-sensor evaluation. }     \label{tab:sensei_config}
    \begin{tabularx}{\textwidth}{CCCC|CCC} \hline
         Shared weights & Sensor encoding & Normalize & Others & \gls{cropharvest} \hspace{1cm} (\acrshort{f1}) & \gls{lfmc}  \hspace{1cm} (\acrshort{r2}) & \gls{pm25}  \hspace{1cm} (\acrshort{r2}) \\ \hline 
                - & -    &  -  &  &  $0.603$ & $0.313$ & ${0.308}$  \\ \hline
                  $\checkmark$ & - & - &   & $0.595$ & $0.281$ & $\mathbf{0.281}$ \\
                  $\checkmark$ & $\checkmark$ & - &   & $\mathbf{0.606}$ & ${0.309}$ & $0.274$ \\
                  $\checkmark$ & $\checkmark$ & $\checkmark$ & add  & $0.598$ & $\mathbf{0.317}$ & $0.241$ \\
                  $\checkmark$ & $\checkmark$ & $\checkmark$ & concat & $0.595$ & $0.297$ & $0.196$ \\
                  \hline
    \end{tabularx}
\end{table}

\section{Related work} \label{sec:related}

In the following, we discuss some related works in the multi-sensor \gls{eo} literature.

\paragraph{Multi-sensor learning.} 
Recent research in self-supervised learning and foundational models has placed the usage of multi-sensor data as a basis. 
Most of these works use feature-level fusion with sensor-dedicated encoders to handle sensors with different spatial resolutions.
For instance, in Chen et al. \cite{chen2024novel}, a pure Transformer model is introduced that uses cross-modal contrastive and reconstruction learning as pre-training.
While in OmniSat \cite{astruc2024omnisat}, a similar model is introduced based on contrastive and reconstruction learning but adapted to sensors with different temporal resolutions.
In Guo et al. \cite{guo2024skysense}, Skysense is proposed as a geospatial foundational model through cross-modal contrastive learning and knowledge distillation (via a student-teacher framework).
Additionally, some works compare different sensor fusion strategies for pixel-wise \cite{mena2024search} and image-wise \cite{follath2024multi} crop classification with \gls{sits}.
In any of these models, the \gls{sd} technique can be integrated by just randomly masking out all input features from the multi-sensor data. While for the sensor-invariant component, it can be integrated in any of the sensor-dedicated layers by just sharing some parameters.

\paragraph{Random masking.}
Masking out the input data has been used in the literature as an augmentation technique during training. 
For instance, by randomly dropping time-steps of \gls{sits} \cite{saintefaregarnot2022multi}, features in random modalities \cite{tseng2023lightweight}, or patches in random modalities \cite{astruc2024omnisat,chen2024novel}. 
In OmniSat \cite{astruc2024omnisat}, a masking vector is learned that replaces the masked data, while in Presto \cite{tseng2023lightweight} and Chen et al. \cite{chen2024novel} the masked data is ignored by using masked attention models.
In these works, Transformer models are used with a feature-level fusion strategy.

\paragraph{Sensor invariant.}
The introduced ESensI is directly motivated by Francis \cite{francis2024sensor}, which uses the same layers for different sensor descriptors, including a block that generates all possible sensor permutations (\textit{permutation block}). However, in that work, they just focus on satellite image cloud segmentation.
In Hackstein et al. \cite{hackstein2024exploring}, different ways of using shared layers (\textit{common blocks}) are compared for cross-sensor reconstruction. The results show that shared layers are more effective in the decoder than the encoder, as we used. Besides, shared layers allow reducing the computational resources of the multi-sensor model.

\section{Conclusion} \label{sec:conclusion}

We introduced two methods for multi-sensor modeling with missing sensors, \gls{isd} and \gls{sensei}.
The validation in three time series \gls{eo} datasets showed that our methods can significantly increase the model robustness to missing sensors. 
However, there is a limitation regarding predictive performance when all sensors are available.
Future work will focus on detecting the sources of robustness, as it is unclear if it comes from: (i) the model's adaptability, (ii) model's disregard for that sensor information, or (iii) the irrelevance of the sensor data for prediction.

\begin{credits}
\subsubsection{\ackname} 
F. Mena acknowledges support through a scholarship of the University of Kaiserslautern-Landau. 
\end{credits}

\bibliographystyle{splncs04}
\bibliography{main}

\begin{thebibliography}{10}
\providecommand{\url}[1]{\texttt{#1}}
\providecommand{\urlprefix}{URL }
\providecommand{\doi}[1]{https://doi.org/#1}

\bibitem{aksoy2009land}
Aksoy, S., Koperski, K., Tusk, C., Marchisio, G.: Land cover classification
  with multi-sensor fusion of partly missing data. Photogrammetric Engineering
  \& Remote Sensing  \textbf{75}(5),  577--593 (2009).
  \doi{10.14358/PERS.75.5.577}

\bibitem{astruc2024omnisat}
Astruc, G., Gonthier, N., Mallet, C., Landrieu, L.: {OmniSat}:
  {Self}-supervised modality fusion for {Earth} observation. arXiv preprint
  arXiv:2404.08351  (2024)

\bibitem{pm25}
Chen, S.: {PM2.5 Data of Five Chinese Cities}. UCI Machine Learning Repository
  (2017). \doi{10.24432/C52K58}

\bibitem{chen2024novel}
Chen, Y., Zhao, M., Bruzzone, L.: A novel approach to incomplete multimodal
  learning for remote sensing data fusion. IEEE Transactions on Geoscience and
  Remote Sensing  (2024). \doi{10.1109/TGRS.2024.3387837}

\bibitem{efremova2021soil}
Efremova, N., Seddik, M.E.A., Erten, E.: Soil moisture estimation using
  {Sentinel-1/-2} imagery coupled with {cycleGAN} for time-series gap filing.
  IEEE Transactions on Geoscience and Remote Sensing  \textbf{60},  1--11
  (2021). \doi{10.1109/TGRS.2021.3134127}

\bibitem{follath2024multi}
Follath, T., Mickisch, D., Hemmerling, J., Erasmi, S., Schwieder, M., Demir,
  B.: Multi-modal vision transformers for crop mapping from satellite image
  time series. In: Proceedings of IEEE {International Geoscience} and {Remote
  Sensing Symposium} (2024)

\bibitem{francis2024sensor}
Francis, A.: Sensor independent cloud and shadow masking with partial labels
  and multimodal inputs. IEEE Transactions on Geoscience and Remote Sensing
  (2024). \doi{10.1109/TGRS.2024.3391625}

\bibitem{guo2024skysense}
Guo, X., Lao, J., Dang, B., Zhang, Y., Yu, L., Ru, L., Zhong, L., Huang, Z.,
  Wu, K., Hu, D., et~al.: Skysense: {A} multi-modal remote sensing foundation
  model towards universal interpretation for {Earth} observation imagery. In:
  Proceedings of IEEE/CVF Conference on Computer Vision and Pattern
  Recognition. pp. 27672--27683 (2024)

\bibitem{hackstein2024exploring}
Hackstein, J., Sumbul, G., Clasen, K.N., Demir, B.: Exploring masked
  autoencoders for sensor-agnostic image retrieval in remote sensing. arXiv
  preprint arXiv:2401.07782  (2024)

\bibitem{heinrich2023targeted}
Heinrich, R., Scholz, C., Vogt, S., Lehna, M.: Targeted adversarial attacks on
  wind power forecasts. Machine Learning  (2023).
  \doi{10.1007/s10994-023-06396-9}

\bibitem{hong2021more}
Hong, D., Gao, L., Yokoya, N., Yao, J., Chanussot, J., Du, Q., Zhang, B.: More
  diverse means better: {Multimodal} deep learning meets remote-sensing imagery
  classification. IEEE Transactions on Geoscience and Remote Sensing
  \textbf{59},  4340--4354 (2021). \doi{10.1109/TGRS.2020.3016820}

\bibitem{ko2008dynamic}
Ko, A.H., Sabourin, R., Britto~Jr, A.S.: From dynamic classifier selection to
  dynamic ensemble selection. Pattern recognition  \textbf{41}(5),  1718--1731
  (2008). \doi{10.1016/j.patcog.2007.10.015}

\bibitem{little2019statistical}
Little, R.J., Rubin, D.B.: Statistical analysis with missing data. John Wiley
  \& Sons (2002)

\bibitem{ma2022multimodal}
Ma, M., Ren, J., Zhao, L., Testuggine, D., Peng, X.: Are multimodal
  transformers robust to missing modality? In: Proceedings of IEEE/CVF
  Conference on Computer Vision and Pattern Recognition. pp. 18177--18186
  (2022). \doi{10.1109/CVPR52688.2022.01764}

\bibitem{mena2024igarss}
Mena, F., Arenas, D., Charfuelan, M., Nuske, M., Dengel, A.: Impact assessment
  of missing data in model predictions for {Earth} observation applications.
  In: Proceedings of IEEE {International Geoscience} and {Remote Sensing
  Symposium} (2024)

\bibitem{mena2024search}
Mena, F., Arenas, D., Dengel, A.: In the search for optimal multi-view learning
  models for crop classification with global remote sensing data. arXiv
  preprint arXiv:2403.16582  (2024)

\bibitem{mena2024common}
Mena, F., Arenas, D., Nuske, M., Dengel, A.: Common practices and taxonomy in
  deep multi-view fusion for remote sensing applications. IEEE Journal of
  Selected Topics in Applied Earth Observations and Remote Sensing pp. 4797 --
  4818 (2024). \doi{10.1109/JSTARS.2024.3361556}

\bibitem{pathak2023predicting}
Pathak, D., Miranda, M., Mena, F., Sanchez, C., Helber, P., Bischke, B.,
  et~al.: Predicting crop yield with machine learning: {An} extensive analysis
  of input modalities and models on a field and sub-field level. In:
  Proceedings of IEEE International Geoscience and Remote Sensing Symposium.
  pp. 2767--2770 (2023). \doi{10.1109/IGARSS52108.2023.10282318}

\bibitem{rao2020sar}
Rao, K., Williams, A.P., Flefil, J.F., Konings, A.G.: {SAR}-enhanced mapping of
  live fuel moisture content. Remote Sensing of Environment  \textbf{245},
  111797 (2020). \doi{10.1016/j.rse.2020.111797}

\bibitem{saintefaregarnot2022multi}
Sainte Fare~Garnot, V., Landrieu, L., Chehata, N.: Multi-modal temporal
  attention models for crop mapping from satellite time series. ISPRS Journal
  of Photogrammetry and Remote Sensing  \textbf{187},  294--305 (2022).
  \doi{10.1016/j.isprsjprs.2022.03.012}

\bibitem{scarpa2018cnn}
Scarpa, G., Gargiulo, M., Mazza, A., Gaetano, R.: A {CNN}-based fusion method
  for feature extraction from {Sentinel} data. Remote Sensing  \textbf{10}(2),
  ~236 (2018). \doi{10.3390/rs10020236}

\bibitem{srivastava2012multimodal}
Srivastava, N., Salakhutdinov, R.R.: Multimodal learning with deep {Boltzmann}
  machines. Advances in Neural Information Processing  \textbf{25} (2012)

\bibitem{srivastava2019understanding}
Srivastava, S., Vargas-Munoz, J.E., Tuia, D.: Understanding urban landuse from
  the above and ground perspectives: {A} deep learning, multimodal solution.
  Remote Sensing of Environment  \textbf{228},  129--143 (2019).
  \doi{10.1016/j.rse.2019.04.014}

\bibitem{tseng2023lightweight}
Tseng, G., Cartuyvels, R., Zvonkov, I., Purohit, M., Rolnick, D., Kerner, H.:
  Lightweight, pre-trained transformers for remote sensing timeseries (2023)

\bibitem{tseng2021crop}
Tseng, G., Zvonkov, I., Nakalembe, C.L., Kerner, H.: {{CropHarvest}}: {{A}}
  global dataset for crop-type classification. Proceedings of NIPS Datasets and
  Benchmarks  (2021)

\end{thebibliography}

\clearpage
\appendix

\textsc{\large Supplementary Material}

\section{Dataset setting}
In Table~\ref{tab:datasets} we display the number of features per sensor in each dataset. Regarding the data preprocessing, we apply a z-score normalization to the input data and encode the categorical and ordinal features (like land-cover and canopy height) with a one-hot-vector.
\begin{table}[!h]
    \centering
    \caption{Number of features per sensor source in each dataset. In parentheses is the percentage of the total number of dimensions (if the time series are flattened). } \label{tab:datasets}
    \begin{tabularx}{\textwidth}{c|CCCCCCC} \hline
         Dataset &  Optical & Radar & Weather & Static & Conditions & Dynamics  & Precipitation \\ \hline
         \acrshort{cropharvest} &  11 ($0.73$) & 2 ($0.13$) & 2 ($0.13$) & 2 ($0.01$)&  \\
         \acrshort{lfmc} & 8 ($0.63$) & 3 ($0.24$) &   & 7 ($0.14$) &   \\
         \acrshort{pm25} &   & & & & 3  ($0.32$) & 4 ($0.45$) & 2 ($0.23$) \\ \hline
    \end{tabularx}
\end{table}

\section{Predictive performance results}

\begin{figure}[!b]
    \centering
    \includegraphics[width=0.99\textwidth]{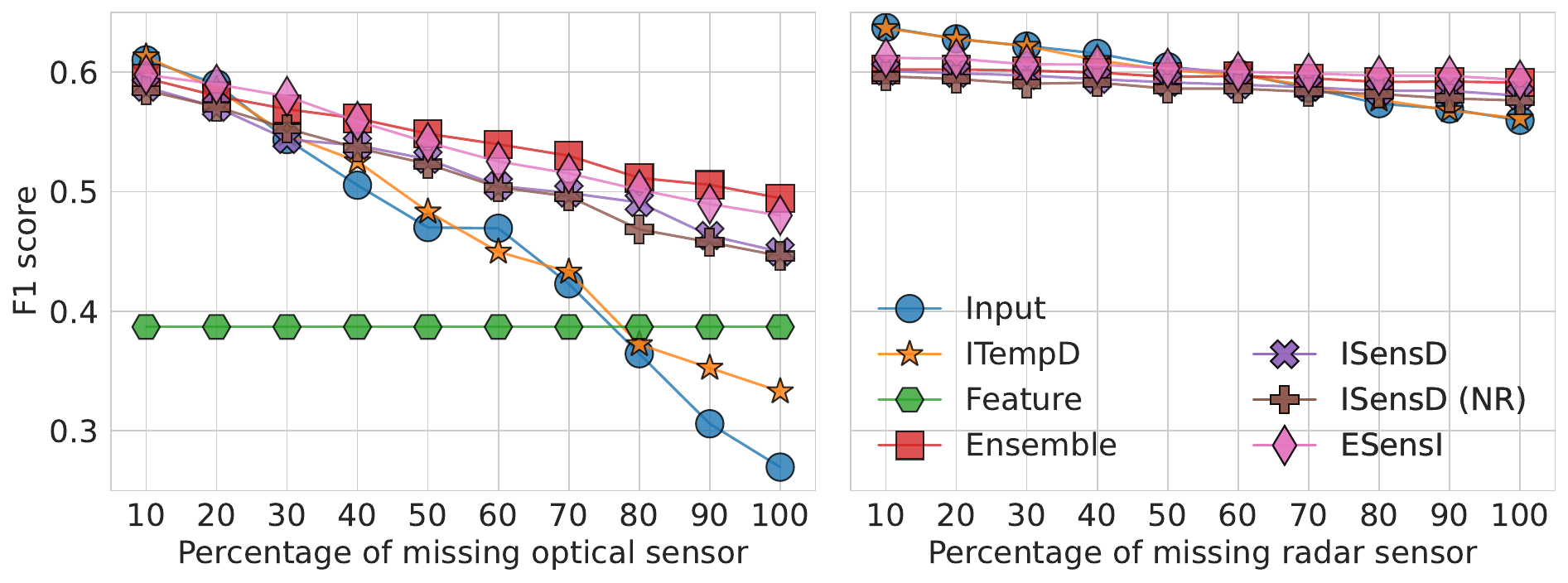}
    \caption{\gls{f1} scores at different levels of missing sensors in the \gls{cropharvest} data.} \label{fig:app:cropharvest_results}
\end{figure}

\begin{figure}[!t]
    \centering
    \includegraphics[width=0.99\textwidth]{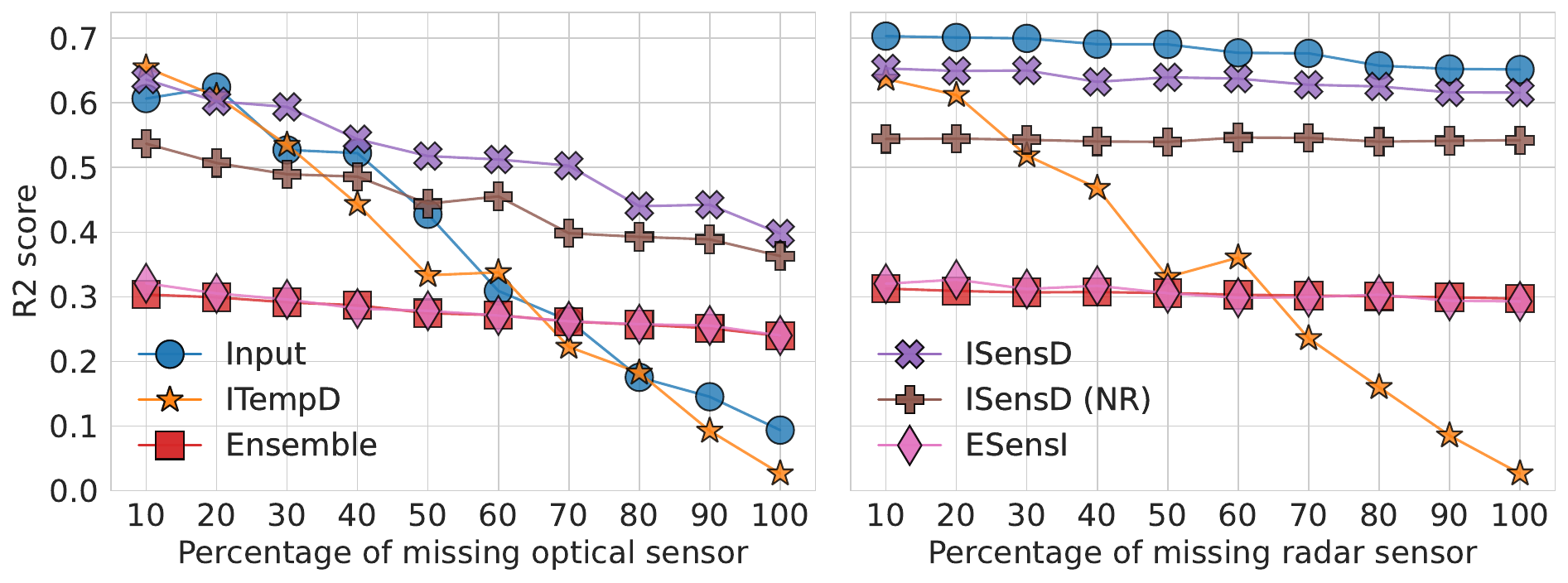}
    \caption{\gls{r2} scores of the compared methods at different levels of missing sensors in the \gls{lfmc} data.} \label{fig:app:lfmc_results}
\end{figure}

\begin{figure}[!t]
    \centering
    \includegraphics[width=0.99\textwidth]{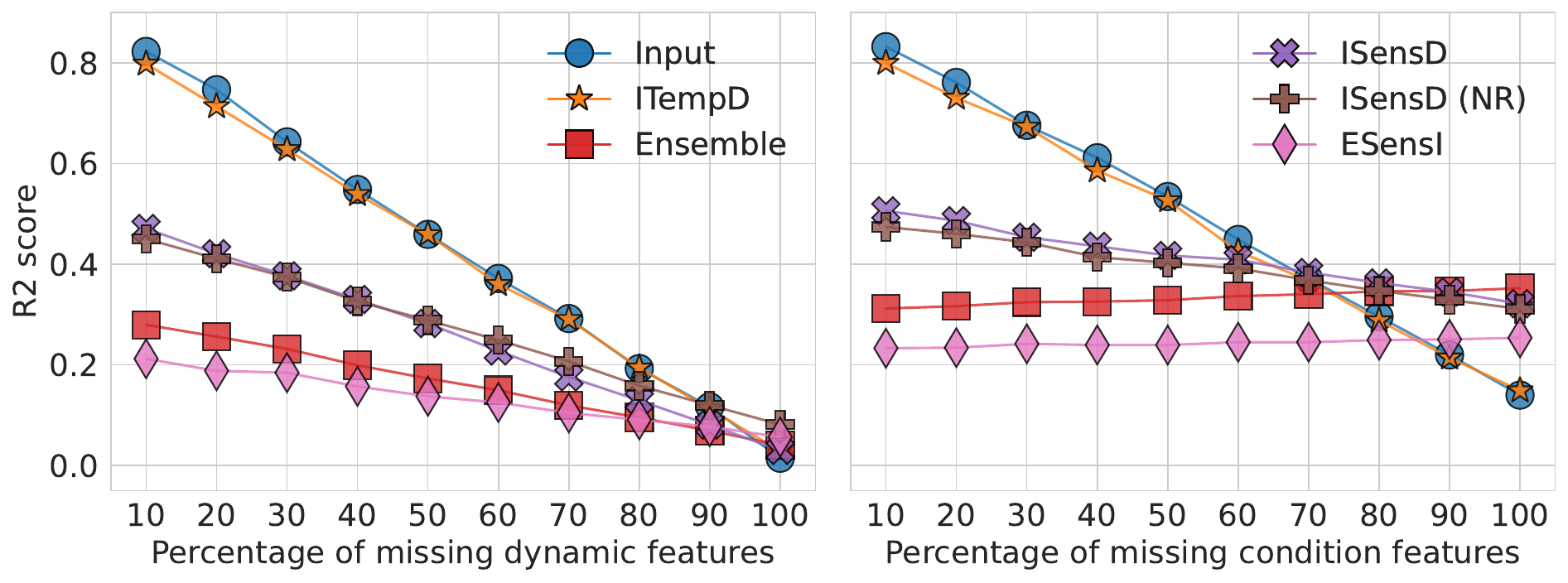}
    \caption{\gls{r2} scores at different levels of missing sensors in the \gls{pm25} data.} \label{fig:app:pm25_results}
\end{figure}

As a complement to the \gls{prs} results shown in the main content of the manuscript, we include predictive performance scores, the \gls{f1} in classification, and \gls{r2} in regression. Fig.~\ref{fig:app:cropharvest_results}, \ref{fig:app:lfmc_results}, \ref{fig:app:pm25_results} show the results for the \gls{cropharvest}, \gls{lfmc}, and \gls{pm25} datasets respectively.
In classification, the ensemble-based methods are the best in the \gls{f1} score. However, in regression, the input-level fusion methods are the best in the \gls{r2} score, while for missing sensor cases, the proposed methods with \gls{sd} are the best options.

\section{Alternative encoders}
\begin{figure}[!t]
    \centering
    \subfloat
    {\includegraphics[width=0.95\textwidth]{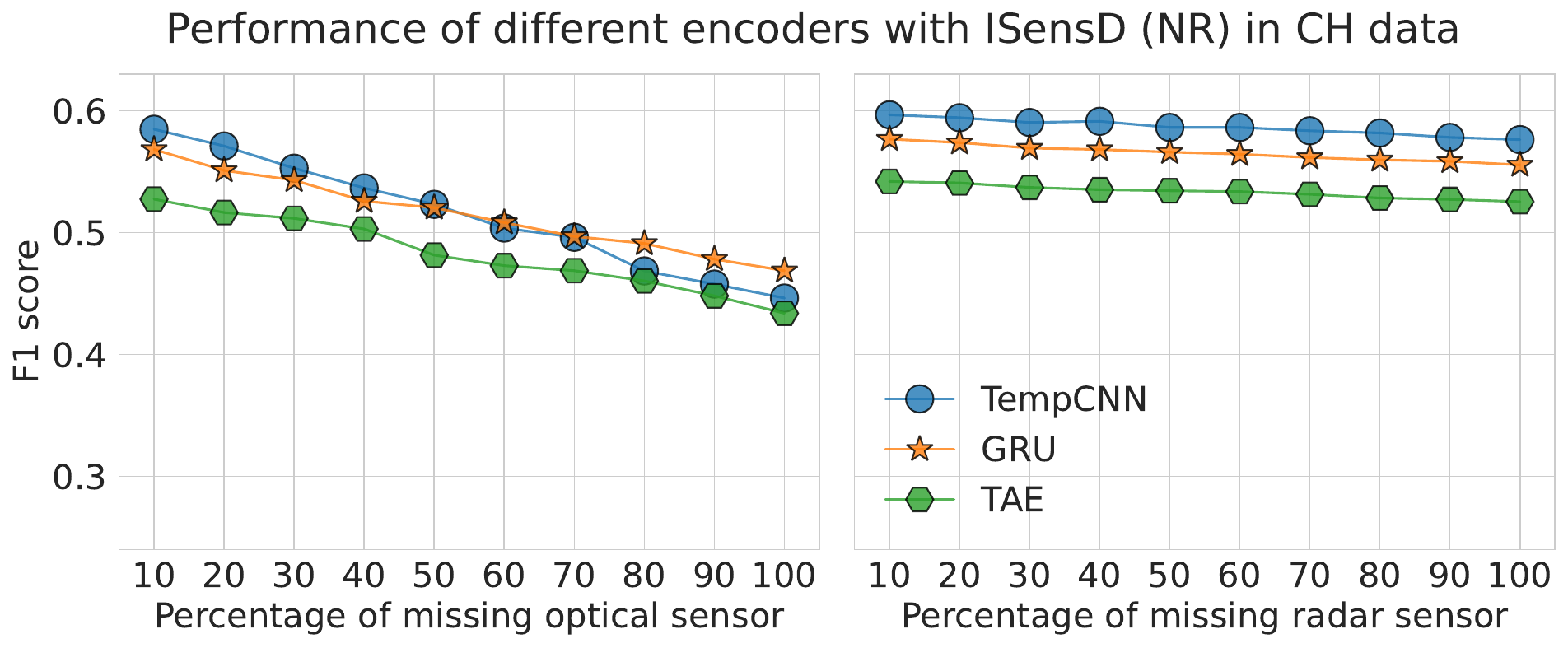}} \\
    \subfloat
    {\includegraphics[width=0.95\textwidth]{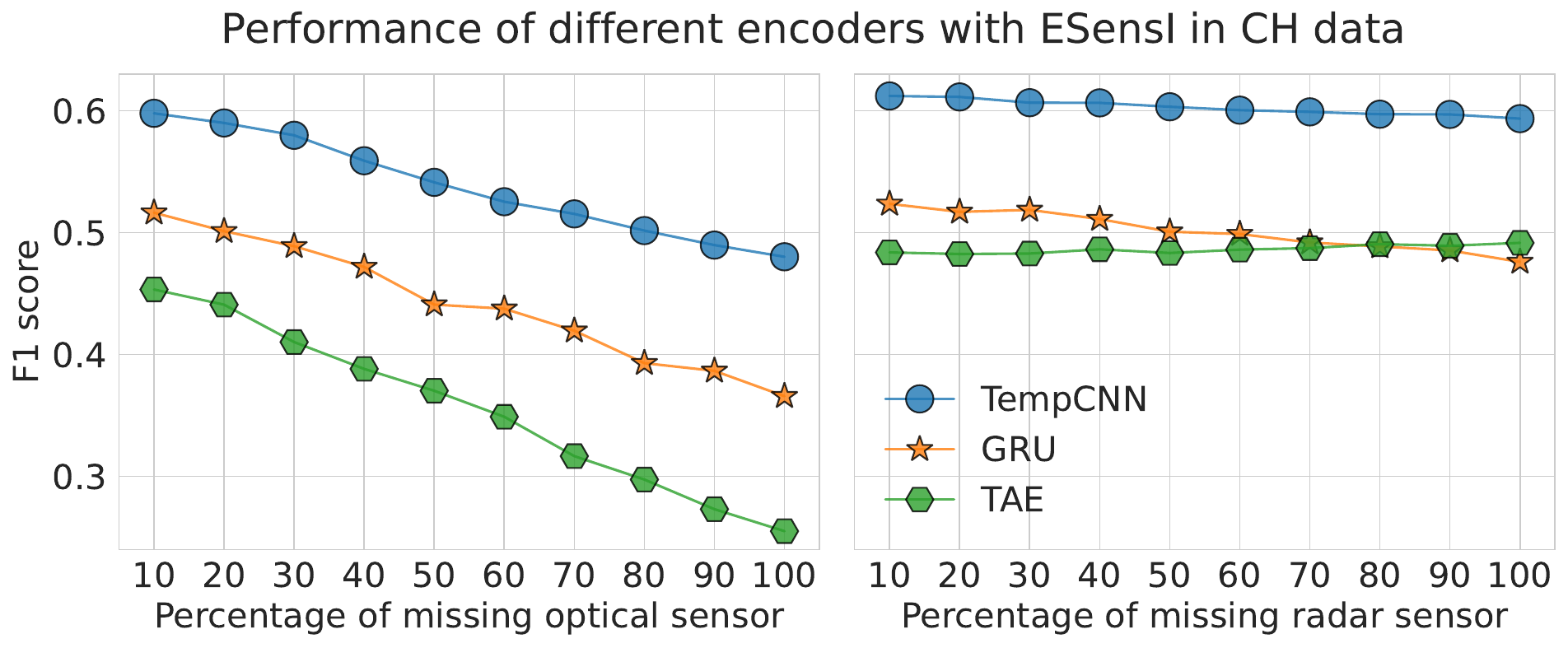}}\\
    \caption{Predictive performance with different encoders in the \gls{cropharvest} data.} \label{fig:encoders:ch}
\end{figure}
\begin{figure}[!t]
    \centering
    \subfloat
    {\includegraphics[width=0.95\textwidth]{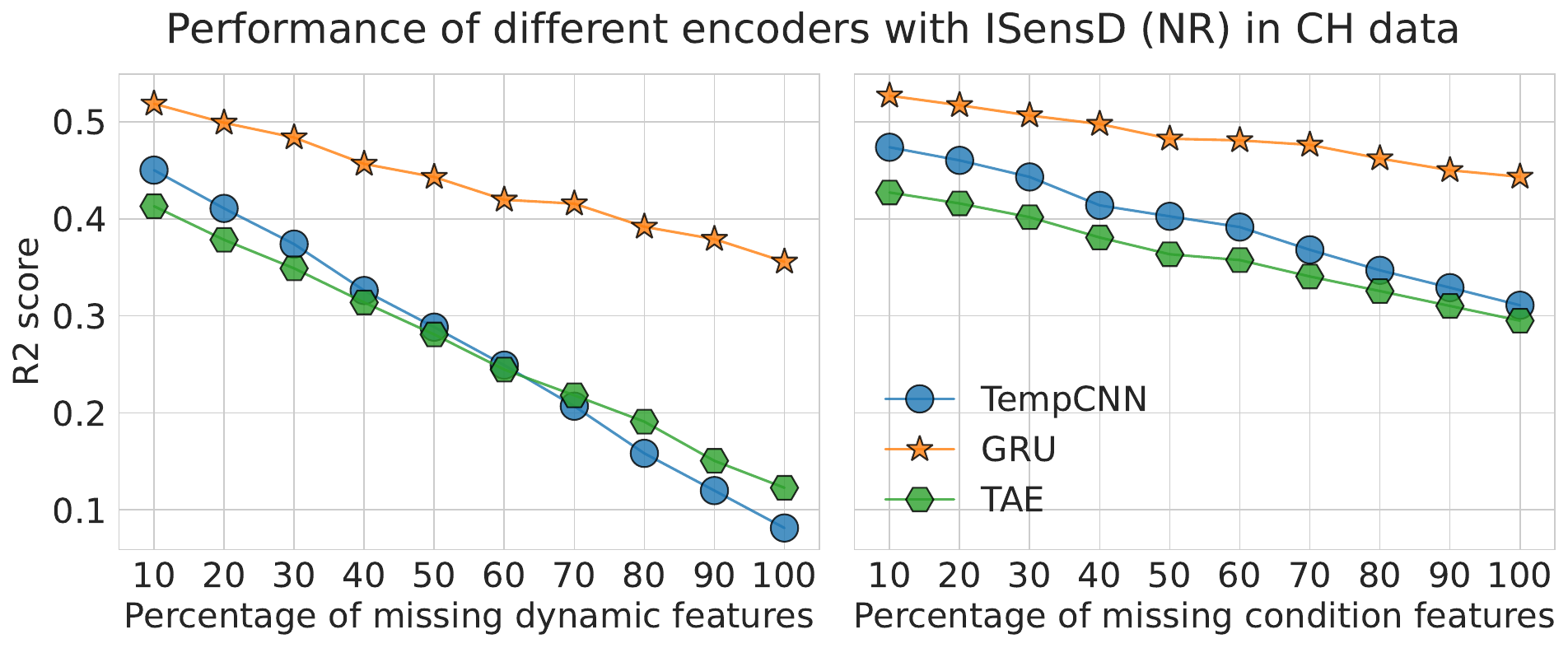}} \\    
    \caption{Predictive performance with different encoders in the \gls{pm25} data.} \label{fig:encoders:pm25}
\end{figure}
\begin{figure}[!t]
    \centering
    \subfloat
    {\includegraphics[width=0.95\textwidth]{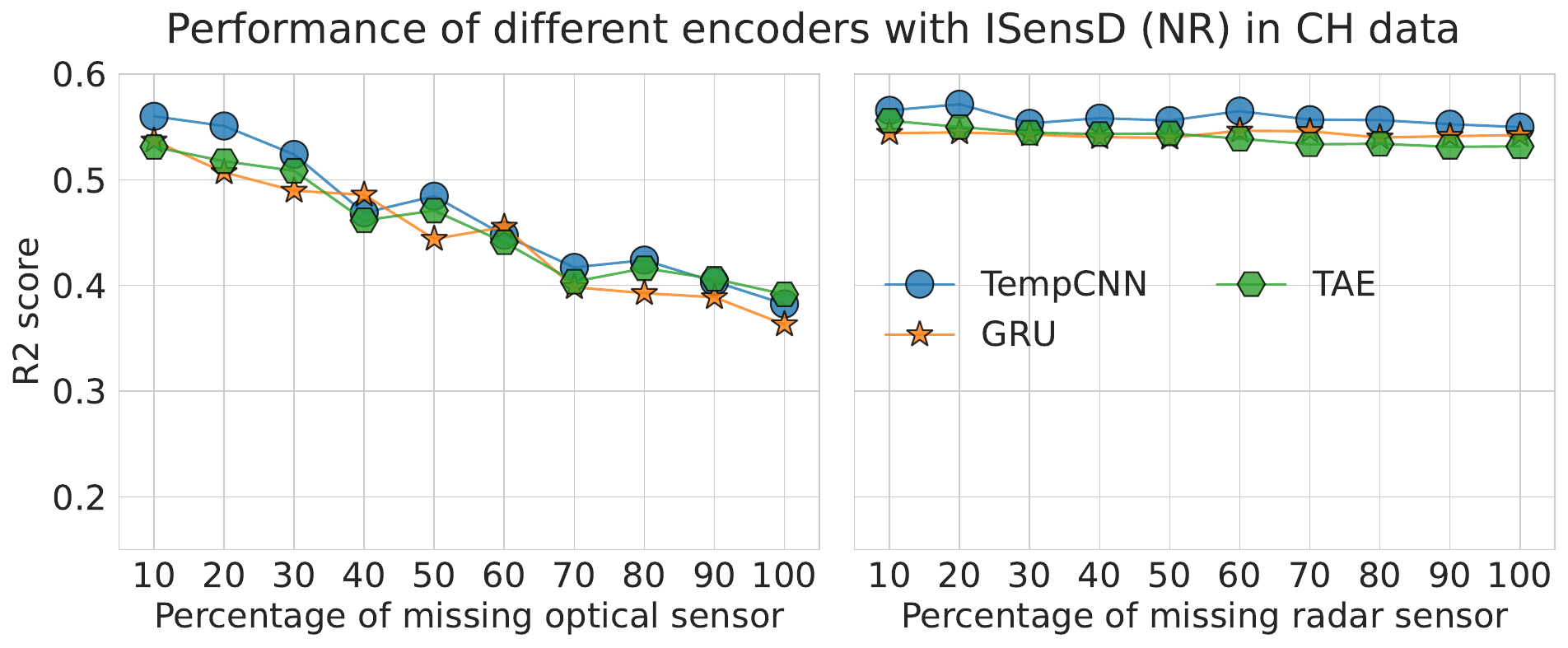}} \\
    \subfloat
    {\includegraphics[width=0.95\textwidth]{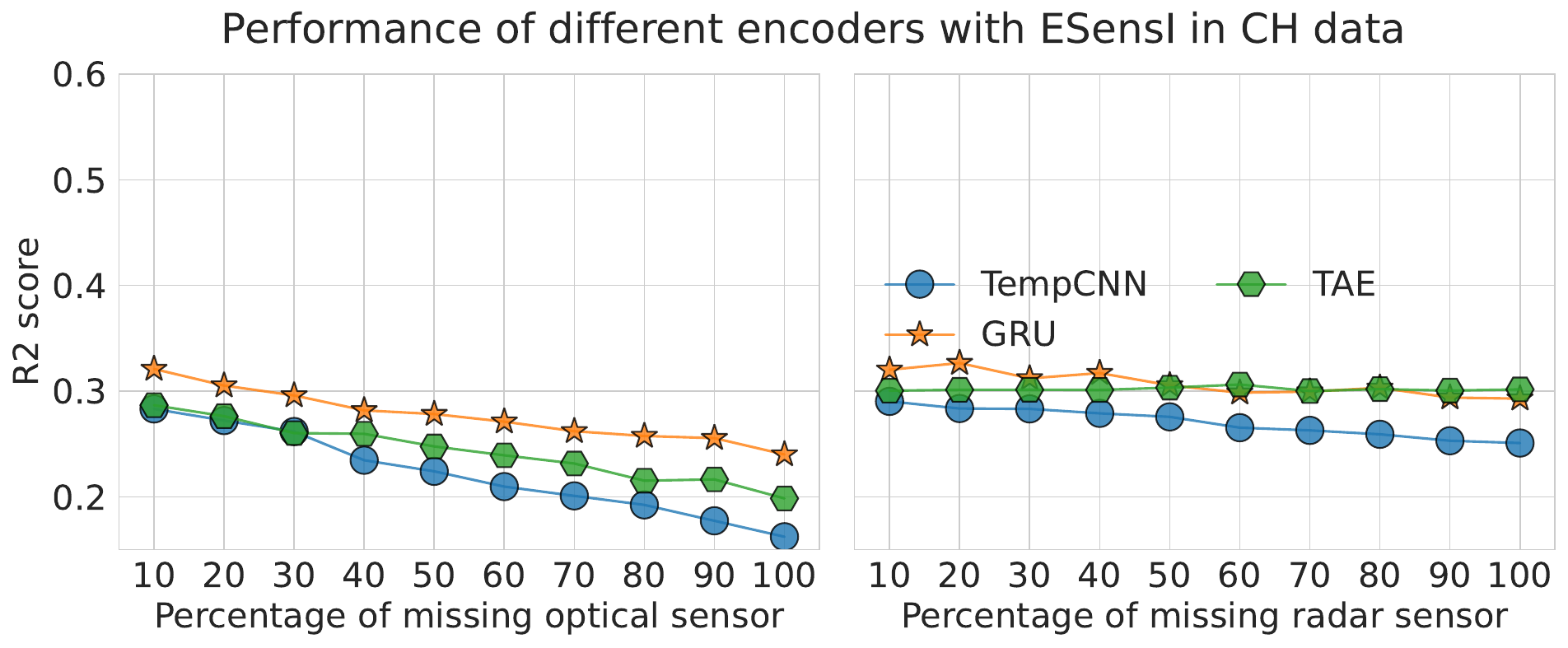}}\\
    \caption{Predictive performance with different encoders in the \gls{lfmc} data.} \label{fig:encoders:lfmc}
\end{figure}
In Fig.~\ref{fig:encoders:ch}, \ref{fig:encoders:pm25}, and  \ref{fig:encoders:lfmc},ex we present the predictive performance results when the encoder architecture in the proposed models is changed. The options explored are TempCNN (the default we use in our paper, 1D CNN applied over time), Temporal Attention Encoder (TAE), and Gated Recurrent Unit (GRU)-based RNN.  
In some cases, the TempCNN architecture is more robust, like in the \gls{cropharvest} data. While in others, the GRU architecture is more robust, like in the \gls{pm25} data.
However, we can see that the main pattern is the same for all encoder architectures, with no extreme changes in the overall results.

\end{document}